\documentclass[11pt]{article}

\usepackage{acl}

\usepackage{times}
\usepackage{latexsym}

\usepackage{tikz-dependency} 
\usepackage{sdrt} 
\usepackage{subcaption} 
\usepackage{stmaryrd} 
\usepackage{amsmath} 
\usepackage{array} 
\usepackage{multirow} 
\usepackage{comment} 

\newcommand{\tll}{\(\llbracket\)}
\newcommand{\trr}{\(\rrbracket\)}

\newcommand{\obj}{\textit} 
\def\arraystretch{1.2} 

\makeatletter
\providecommand{\@LN}[2]{}
\makeatother

\title{A Compositional Typed Semantics for Universal Dependencies}

\author{Laurestine Bradford, Timothy John O'Donnell, Siva Reddy} 

\begin{document}

\maketitle

\begin{abstract}

Languages may encode similar meanings using different sentence structures. This makes it a challenge to provide a single set of formal rules that can derive meanings from sentences in many languages at once. To overcome the challenge, we can take advantage of language-general connections between meaning and syntax, and build on cross-linguistically parallel syntactic structures. We introduce \emph{UD Type Calculus}, a compositional, principled, and language-independent system of semantic types and logical forms for lexical items which builds on a widely-used language-general dependency syntax framework. We explain the essential features of UD Type Calculus, which all involve giving dependency relations denotations just like those of words. These allow UD-TC to derive correct meanings for sentences with a wide range of syntactic structures by making use of dependency labels. Finally, we present evaluation results on a large existing corpus of sentences and their logical forms, showing that UD-TC can produce meanings comparable with our baseline.

\end{abstract}

\section{Introduction}

The connection between a sentence and its meaning is not always straightforward. Semanticists have developed a number of theoretical tools to turn sentences into unambiguous logical forms representing their meanings. However, it is challenging to provide a single framework that can do so across many languages and grammatical constructions. Such a semantic framework would allow us to extend insights from well-studied languages to less well-studied ones. We therefore pursue such a system in this paper.   

Fortunately, there are systematic language-general links between sentence structure and meaning structure. For example, verbs tend to denote actions, and have specific syntactic configurations with the actions' participants. The present work takes advantage of links like these to develop a semantic framework that is wide-coverage in terms of both linguistic constructions and languages. In particular, we create a compositional semantics based on Universal Dependencies syntax \citep{ud}. This syntactic annotation scheme is designed to create parallel trees for parallel sentences in very different languages, so we can use it to compute logical forms in parallel ways as well.

We present UD Type Calculus, a compositional formal semantic framework for Universal Dependencies. We outline the essential properties of the semantic framework. We then evaluate its coverage computationally by creating a lexicon according to its principles and showing that, with this lexicon, gold meanings in an existing meaning bank are derivable within our framework.

\section{Background}

\subsection{Universal Dependencies}
\begin{figure}
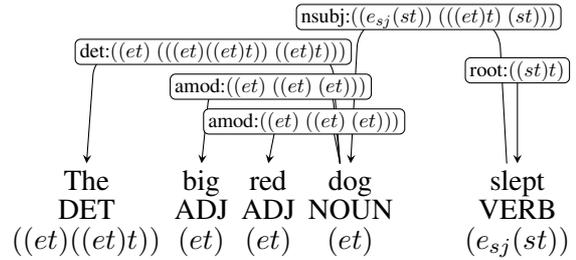

\centering
\begin{dependency}
\begin{deptext}
{The} \& {big} \& {red} \& {dog} \&[2em] {slept}\\
{DET} \& {ADJ} \& {ADJ} \& {NOUN} \& {VERB}\\
\(((et)((et)t))\) \& \((et)\) \& \((et)\) \& \((et)\) \& \((e_{sj}(st))\)\\
\end{deptext}
\depedge[edge height=5em]{5}{4}{nsubj:\(((e_{sj}(st))\ (((et)t)\ (st)))\)}
\depedge[edge height=3.8em]{4}{1}{det:\(((et)\ (((et)((et)t))\ ((et)t)))\)}
\depedge[edge height=2.6em]{4}{2}{amod:\(((et)\ ((et)\ (et)))\)}
\depedge[edge height=1.4em]{4}{3}{amod:\(((et)\ ((et)\ (et)))\)}
\deproot[edge height=3.8em]{5}{root:\(((st)t)\)}
\end{dependency}
\caption{Universal Dependencies syntax (in upright text) and UD-TC semantic types (in italics) for tokens and relations in \emph{The big red dog slept}.\label{fig:bigreddog}}
\end{figure}

In order to create a semantic framework that applies well to many languages, we base it on a syntactic framework designed to annotate syntactic structures consistently across a wide variety of languages. Universal Dependencies \citep[UD;][]{ud} is such a syntactic framework.

UD is a \textit{dependency syntax} framework: the structure of a sentence is expressed as a collection of labelled, directed binary relations between tokens. Taken together, the relations in a sentence form a directed tree, with the main predicate as the root. The relations and the parts-of-speech of words are labeled from fixed, universal lists of labels. Figure \ref{fig:bigreddog} shows an example of a UD tree.
 
There is extensive data annotated with UD syntax, with treebanks in over 140 languages from over 20 families \citep{udtreebanks1,udtreebanks2}.

\subsection{Discourse Representation Structures}

\begin{figure}
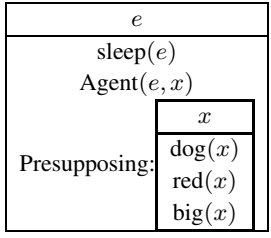

\small
\[\begin{array}{|c|}
\hline
e\\\hline
\textup{sleep}(e)\\
\textup{Agent}(e,x)\\
\textup{Presupposing:} \begin{array}{|c|}\hline x\\\hline
\textup{dog}(x)\\\textup{red}(x)\\\textup{big}(x)\\\hline\end{array}\\\hline\end{array}\]
\caption{The final computed Discourse Representation Structure logical form for \obj{The big red dog slept}. We omit tense information for space.\label{fig:bigreddogmeaning}}
\end{figure}

Due to the availability of cross-linguistic evaluation data, we use \textit{discourse representation structures} \citep[DRS;][]{drt} to represent meanings in logical form.

Each DRS consists of an upper box introducing referents and a lower box introducing assertions made of those referents. We also allow DRS structures to bear labelled discourse relations to each other, as in Segmented Discourse Representation Theory \cite{sdrt}. For example, the DRS in Figure \ref{fig:bigreddogmeaning} introduces a referent and asserts that it is a sleeping event. It presupposes another DRS introducing a referent and asserting that it is big, red, and a dog. The larger DRS asserts that the event's agent is the referent \(x\) from the smaller DRS.

Large meaning banks are available with meanings in DRS format \citep{gmb,pmb}, so this format has been used by a body of previous work on computing meaning from UD \citep{udboxer} and semantic parsing (\citealp{shenandevang, wangetal,sharedtask}).

\section{UD Type Calculus}
\label{sec:binarization}

\begin{figure}
\small \[\llbracket\textup{amod}\rrbracket = \lambda F. \lambda G. \lambda x. F(x) \wedge G(x)\]
\[\llbracket\textup{\obj{dog}}\rrbracket = \lambda x. \begin{array}{|c|}\hline\ \\\hline\textup{dog}(x)\\\hline\end{array}\]
\[\llbracket\textup{\obj{red}}\rrbracket = \lambda x. \begin{array}{|c|}\hline\ \\\hline\textup{red}(x)\\\hline\end{array}\]
\[\llbracket\textup{amod}\rrbracket(\llbracket\textup{\obj{dog}}\rrbracket)(\llbracket\textup{\obj{red}}\rrbracket) = \lambda x. \begin{array}{|c|}\hline\ \\\hline\textup{dog}(x)\\\textup{red}(x)\\\hline\end{array}\]
\caption{Example computation of a phrase denotation by applying \(\llbracket\textup{amod}\rrbracket\) to two word denotations. \label{fig:functionexample}}
\end{figure}

In UD Type Calculus (UD-TC), the meanings, or \textit{denotations}, of words and dependency relations are represented as DRS functions. Relation denotations are two-place functions taking their head's and dependent's denotations as arguments and returning a denotation for the combined phrase. For example, in Figure \ref{fig:functionexample}, we compute a denotation for \obj{red dog} by applying the denotation of ``amod'' (the adjective relation) to those of \obj{dog} (the head) and \obj{red} (the dependent). UD contains one privileged unary relation ``root,'' whose denotation is a one-place function.

\begin{figure}[h]
\centering
\subcaptionbox{Possible composition orders for relation denotations in \obj{the big red dog.}
\label{fig:bigreddoggood}}[0.45\textwidth]{
\small \[\llbracket\textup{det}\rrbracket\Big(\llbracket\textup{amod}\rrbracket\big(\llbracket\textup{amod}\rrbracket(\llbracket\textup{\obj{dog}}\rrbracket)(\llbracket\textup{\obj{red}}\rrbracket)\big)(\llbracket\textup{\obj{big}}\rrbracket)\Big)(\llbracket\textup{\obj{the}}\rrbracket)\]
\[\llbracket\textup{det}\rrbracket\Big(\llbracket\textup{amod}\rrbracket\big(\llbracket\textup{amod}\rrbracket(\llbracket\textup{\obj{dog}}\rrbracket)(\llbracket\textup{\obj{big}}\rrbracket)\big)(\llbracket\textup{\obj{red}}\rrbracket)\Big)(\llbracket\textup{\obj{the}}\rrbracket)\]}

\subcaptionbox{An impossible composition order for relation denotations in \obj{the big red dog.}\label{fig:bigreddogbad}}[0.45\textwidth]{
\small \[\llbracket\textup{amod}\rrbracket\Big(\llbracket\textup{det}\rrbracket\big(\llbracket\textup{amod}\rrbracket(\llbracket\textup{\obj{dog}}\rrbracket)(\llbracket\textup{\obj{red}}\rrbracket)\big)(\llbracket\textup{\obj{the}}\rrbracket)\Big)(\llbracket\textup{\obj{big}}\rrbracket)\]}

\subcaptionbox{The only possible composition order for ``nsubj'' and ``root'' relation denotations in \obj{The big red dog slept.} \label{fig:nsubjslept}}[0.45\textwidth]{
\small \[\llbracket\textup{root}\rrbracket\big(\llbracket\textup{nsubj}\rrbracket(\llbracket\textup{\obj{slept}}\rrbracket)(\llbracket\textup{\obj{the big red dog}}\rrbracket)\big)\]}
\caption{Possible and impossible composition orders for functions denoted by relations in \obj{The big red dog slept}.}
\end{figure}

When multiple relations have the same head, the output from one relation's function serves as input to another. We may compose these functions in multiple orders. For example, the phrase \obj{the big red dog} in Figure \ref{fig:bigreddog} has three relations headed by \emph{dog}. As in Figure \ref{fig:bigreddoggood}, the denotation of the phrase may be computed by applying the denotation of ``amod'' to \tll \obj{dog}\trr\ and \tll \obj{red}\trr\ first, and then treating this combined phrase as the head of the next ``amod''; or, we may apply the denotation of ``amod'' to \tll \obj{dog}\trr\ and \tll \obj{big}\trr\ first, and then treat this combined phrase as the head of the other ``amod.'' Function composition order can sometimes, albeit rarely, affect the overall denotation of a phrase. (See Appendix \ref{app:workedexamples} for an example involving quantifier scope.)

After applying the functions denoted by relations with a given head, the output is passed as an argument to the next higher relation in the tree. For example, in Figure \ref{fig:bigreddog}, we apply the denotations of ``det,'' ``amod,'' and ``amod'' to \tll \obj{dog}\trr\ to create a denotation for \obj{the big red dog}. Only after that do we pass this new denotation as the second argument to the relation ``nsubj,'' as shown in Figure \ref{fig:nsubjslept}.

In addition to denotations, relations and words are given \textit{semantic types}, which constrain the order of composition of functions. A semantic type can be either atomic or composite. An atomic type may be \(e\) (for entity), \(s\) (for event), or \(t\) (for a DRS).\footnote{Other semantic frameworks use \(t\) to represent the type of a truth value, but in these frameworks a sentence's meaning is generally either a truth value or a function from possible worlds to truth values. We assume the meaning of a sentence is a DRS, which may be true or false in a given world, but which also has additional structure.} Type \(e\) has three syntactically-tagged sub-types, \(e_{sj}\) (subject), \(e_{oj}\) (object), and \(e_{io}\) (indirect object); we use these in relation denotations to specify the syntactic role which can saturate each argument. A composite type is the type of a function. It consists of an ordered pair \((xy)\) of two types \(x\) and \(y\), each of which may be atomic or composite. Then \(x\) represents the type of input allowed by a function and \(y\) represents the type of output produced by that function. For example, a function of type \((e_{sj}(st))\) takes a single subject entity as input, and returns a function from events to DRS structures. An intransitive verb such as \obj{slept} in Figure \ref{fig:bigreddog} has this type: the entity argument is the single participant in the action, and the event argument is the action itself. Semantic types can rule out potential composition orders of relation denotations. Figure \ref{fig:bigreddogbad} shows an example impossible composition order: The output of the function \tll det\trr\ is of type \(((et)t)\) and is unsuitable as an input to the function \tll amod\trr.

A single UD tree is often compatible with many possible logical forms \citep{gothamandhaug}, even when constrained by semantic types. When multiple composition orders of relations are possible, we treat them all as possible outputs of the system, thereby sometimes deriving several logical forms for one Universal Dependencies tree.

The UD-TC framework consists entirely of the rules outlined above, which form an interface between Universal Dependencies and the space of DRS logical forms. It does not specify a lexicon of denotations and semantic types. That said, we carefully selected a type system compatible with those used by many formal semanticists (e.g. \citealt{heimandkratzer}). In the lexicon we use below to test UD-TC computationally, we follow approaches to verb and determiner valency advocated in formal theory (e.g. \citealt{barwiseandcooper}).

\section{Related Work}
\label{sec:relatedwork}

This work builds on \textit{UDepLambda} \citep{reddy-etal-2016-transforming,reddy-etal-2017-universal}. UDepLambda also derives logical forms from UD trees compositionally, using edge meanings as to combine word meanings. However, in UD-TC, unlike in UDepLambda, words can have different semantic types. Using these, we can determine all sensible composition orders, even when multiple orders are possible. UDepLambda uses a rigid obliqueness hierarchy for this instead. In addition, we evaluate UD-TC on an existing meaning bank directly, while UDepLambda is evaluated on a downstream natural language understanding task.

Recent work by \citet{udboxer}, called \textit{UD-Boxer}, also derives DRS structures from UD trees. To do this, it applies graph transformations to a UD tree to create a graph representation of a DRS. Thus, it does not assign a denotation to any individual word or relation. Using UD-TC, by contrast, one can easiliy provide denotations in terms of DRS structures for each word of a sentence. Moreover, UD-Boxer only derives a single logical form for each input sentence, while we compute every possible logical form.

\citet{gothamandhaug} present a compositional semantics for UD which also assigns denotations and semantic types to individual words. However, to combine word meanings, \citeauthor{gothamandhaug} employ \textit{meaning constructors}. These specify syntactic environments in which they apply, then combine word denotations into phrase denotations, sometimes also adding nodes to the syntax tree. In UD-TC, these meaning constructors are replaced with typed relation denotations just like those of words, allowing an elegant treatment of words and relations within lambda calculus. 
In addition, while \citeauthor{gothamandhaug} explain their framework theoretically, we also evaluate on a corpus.

See Appendix \ref{ccgappendix} for a comparison with Combinatory Categorial Grammar.
 
\section{Evaluation}


\subsection{Dataset}

We evaluated on gold data from the Parallel Meaning Bank (PMB; \citealp{pmb}). This is a dataset of over 16,000 sentences across four languages: English, German, Italian, and Dutch. Each data point is has a DRS logical form computed by a parser and corrected by humans. We used Stanza \cite{stanza} to parse the raw text of a PMB data point into UD format, then used UD-TC to compute logical forms of each data point.

\subsection{Lexicon}

To evaluate our overall approach, we hand-crafted a lexicon of word and relation denotations. To assign word denotations, each part-of-speech was associated with a few possible denotation templates. Each template was a lambda expression on DRS structures, possibly containing a blank. Filling the blank with a word's lemma produced the word's denotation. Appendix \ref{app:theorysection} outlines the linguistic choices made in writing these templates.

\subsection{Evaluation Metrics}
\label{sec:evalmetrics}

We used Counter \citep{counter} to evaluate DRS match. Counter measures how well the clauses of the computed DRS match the clauses of a target DRS. After a variable alignment step, it returns an F1 score between these two sets of clauses.

PMB logical forms contain specific argument role labels like Agent and Theme. They also contain numerical indicators called \textit{synsets} that disambiguate word senses by mapping to definitions in WordNet \citep{wordnet}. Word sense disambiguation and argument role labelling are not main goals of either UD-TC or our baseline, so we include a Counter score which ignores this information. PMB logical forms also contain discourse relations, like ``Presupposing'' in Figure \ref{fig:bigreddogmeaning}. Our baseline deliberately ignores some such relations, so as a more fair comparison, we include a modified Counter score which ignores all discourse relations.

We aimed to show that our simple, compositional framework derives a sufficiently general space of meanings to capture the correct logical form. This, even though it also derives others. Therefore, we report only the best score out of our computed meanings for each data point from PMB.

\subsection{Baseline}

We compare UD-TC to UD-Boxer \cite{udboxerthesis}.\footnote{UD-Boxer \url{https://github.com/WPoelman/ud-boxer}}
Like UD-TC, UD-Boxer converts UD trees to DRS meanings with no language-specific dictionary.

\subsection{Results}

\begin{table}
\footnotesize
\centering
{
\def\arraystretch{1}
\begin{tabular}{|>{\centering}p{0.15cm}c|cccc| c|}
\hline
& & \multicolumn{4}{c|}{Evaluation} & \\
& & +D  & +D & -D  & -D & \%\\
& & +L  & -L & +L  & -L & Error\\\hline
\multirow{2}{*}{de} & UD-TC & \textbf{46.9}  & \textbf{53.8} & 51.6  & 60.8 & 0.3\\
& UD-B & 22.9 & 26.6 & \textbf{52.7}  & \textbf{62.0} & 0.1\\\hline
\multirow{2}{*}{en} & UD-TC & \textbf{52.9} & \textbf{62.4} & 63.2 & \textbf{75.4} & 1.4\\
& UD-B & 29.1 & 33.9 & \textbf{64.1}  & 74.7 & 0.4\\\hline
\multirow{2}{*}{it} & UD-TC & \textbf{46.3} & \textbf{52.7} & 47.0  & 55.7 & 0.4\\
& UD-B & 18.3  & 21.0 & \textbf{52.0} & \textbf{60.2} & 0\\\hline
\multirow{2}{*}{nl} & UD-TC & \textbf{41.1}  & \textbf{48.1} & 47.7  & 57.0 & 0.5\\
 & UD-B & 22.0  & 25.8 & \textbf{48.4}  & \textbf{57.9} & 0\\\hline
\end{tabular}
}
\def\arraystretch{1.2}

\caption{Mean F1 score computed by Counter for UD-TC and UD-Boxer for each language and evaluation metric. ``+D'' indicates that discourse relations were taken into account, and ``+L'' that synset and argument role labels were taken into account. ``UD-B'' refers to UD-Boxer. ``\% Error'' is the percentage of data points for which no output was produced. Subcorpora: ``de'' = German, ``en'' = English, ``it'' = Italian, ``nl'' = Dutch.}
\label{tab:mainresults}
\end{table}

Table \ref{tab:mainresults} shows the mean F1 score for each
parsing process, subcorpus, and evaluation. Means omit data points for which no output was produced, with the percentage of such data points in the final column of the table.

UD-TC's best output, on average, strongly outperforms UD-Boxer's sole output in evaluations which include discourse relations. UD-TC also slightly outperforms UD-Boxer on the English subcorpus when discourse relations and word sense and semantic role labelling are all ignored. UD-TC's score is slightly below UD-Boxer's on other evaluation metrics ignoring discourse relations, by between 0.7 and 5 percentage points.

UD-TC fails to produce output for between 0.3\% and 1.4\% of data points in each subcorpus, while for UD-Boxer this is between 0\% and 0.4\%.

\subsection{Discussion}

Our experiment shows that UD-TC can generate logical forms comparable with an existing baseline when evaluated on a cross-linguistic meaning bank.

UD-TC and UD-Boxer differ sharply when evaluations include discourse relations because UD-Boxer does not compute presupposition relations between DRS boxes, while UD-TC does. Presuppositions are frequent in the PMB, occurring in 83\% of all gold logical forms.

We computed every possible logical form of an input sentence, which can take a long time for sentences containing many ambiguous relations. When accounting for all ambiguities, each datapoint produced a median of 154 logical forms, but for a few sentences there were over a million.\footnote{This includes spurious ambiguity. For example, when two binarizations produce the same logical form, this is treated as an ambiguity.} Resource constraints required us to restrict the computation time of each data point, which caused 72\% of the cases where UD-TC produced no output.

\section{Conclusion and Future Work}
We have provided a compositional, typed, cross-linguistically parallel semantics for Universal Dependencies. We have also demonstrated that it can generate logical forms comparable with another UD-based meaning computation process. Our approach is flexible enough to derive several possible logical forms for each tree.

While we created a lexicon by hand for the present evaluation, in the future we hope to use UD-TC in a probabilistic model inducing a semantic lexicon from a corpus, as \citet{abendetal} have shown possible using Combinatory Categorial Grammar. A UD-TC backbone will allow such a model to take advantage of the wealth of UD data available even for understudied languages.

\section{Code Availability}
All code and data necessary to reproduce this project is available on Github at \url{https://github.com/McGill-NLP/ud-to-meaning}.

\section{References}\label{sec:reference}
\bibliography{anthology,eval1,eval1-resources}
\bibliographystyle{acl_natbib}

\appendix

\section{Worked Examples}
\label{app:workedexamples}
Figures \ref{fig:relativeclause} and \ref{fig:scopeambiguity} show examples of selected steps in the computation of meaning for sentences with complicated predicate-argument relationships. Figure \ref{fig:relativeclause} shows the computation of the meaning of subject and object relative clauses. Through type flexibility of ``nsubj'' and ``obj'' relations, we are able to assign the correct argument role to the entity representing the cat in each case, despite the use of the same UD relation ``acl'' for both structures. Figure \ref{fig:scopeambiguity} shows that using ambiguity in binarization order, we derive both possible semantic scopes for sentences involving two quantificational determiners.

\begin{figure*}
\centering
\subcaptionbox{Semantic type and denotation of ``acl'' relation. Recall that \(e_{sj}\) and \(e_{ob}\) are sub-types of \(e\), so that \tll acl\trr\ is agnostic as to what empty argument slot its dependent verb has.}[\textwidth]{
\begin{tabular}{lrr}
\tll acl\trr:&\(((et)\ ((e(st))\ (et)))\)&\(\lambda F. \lambda G. \lambda x. \exists e.[F(x) \wedge G(x,e)]\)
\end{tabular}
}

\vspace{2em}

\subcaptionbox{Universal Dependencies syntax and semantic types for a subject relative clause. We regard \obj{that} as semantically vacuous.}[\textwidth]{
\begin{dependency}
\begin{deptext}
cat \& that \& chased \& a \& mouse\\
NOUN \& PRON \& VERB \& DET \& NOUN\\
\((et)\) \& - \& \((e_{ob}(e_{sj}(st)))\) \& \(((et)((et)t))\) \& \((et)\)\\
\end{deptext}
\depedge[edge height=4em]{1}{3}{acl: \(((et)\ ((e(st))\ (et)))\)}
\depedge{3}{2}{nsubj: -}
\depedge{3}{5}{obj: \(((e_{oj}(e_{sj}(st)))\ (((et)t)\ (e_{sj}(st))))\)}
\depedge{5}{4}{det: \(((et)\ (((et)((et)t))\ ((et)t)))\)}
\end{dependency}}

\vspace{2em}

\subcaptionbox{Selected steps in computation of the logical form of the subject relative clause. We omit tense information for space.\label{fig:subjrelclause}}[\textwidth]{
\begin{tabular}{>{\centering\let\newline\\\arraybackslash\hspace{0pt}}p{6cm}>{\centering\let\newline\\\arraybackslash\hspace{0pt}}p{4cm}>{\centering\let\newline\\\arraybackslash\hspace{0pt}}p{5.5cm}}
\tll \obj{chased a mouse}\trr & \tll\obj{cat}\trr & \tll \obj{cat that chased a mouse}\trr \newline = \tll acl\trr \(\big(\)\tll\obj{cat}\trr\(\big)\big(\)\tll \obj{chased a mouse}\trr\(\big)\) \\
\((e_{sj}(st))\) & \((et)\) & \((et)\)\\
\(\lambda x. \lambda e.
\begin{array}{|c|}\hline
y\\\hline
\textup{chase}(e)\\
\textup{Theme}(e,y)\\
\textup{Agent}(e,x)\\
\textup{mouse}(y)
\\\hline\end{array}
\) & \(
\lambda x.
\begin{array}{|c|}\hline
\\\hline
\textup{cat}(x)
\\\hline\end{array}
\) & \(
\lambda x.
\begin{array}{|c|}\hline
y\ e\\\hline
\textup{cat}(x)\\
\textup{chase}(e)\\
\textup{Theme}(e,y)\\
\textup{Agent}(e,x)\\
\textup{mouse}(y)
\\\hline\end{array}
\)
\end{tabular}
}

\vspace{2em}

\subcaptionbox{Universal Dependencies syntax and semantic types for an object relative clause.}[\textwidth]{
\begin{dependency}
\begin{deptext}
cat \& a \& mouse \& chased\\
NOUN \& DET \& NOUN \& VERB\\
\((et)\) \& \(((et)((et)t))\) \& \((et)\) \& \((e_{ob}(e_{sj}(st)))\) \\
\end{deptext}
\depedge{1}{4}{acl: \(((et)\ ((e(st))\ (et)))\)}
\depedge[edge height=2.5em]{4}{3}{nsubj: \(((e_{oj}(e_{sj}(st)))\ (((et)t)\ (e_{oj}(st))))\)}
\depedge{3}{2}{det: \(((et)\ (((et)((et)t))\ ((et)t)))\)}
\end{dependency}
}

\vspace{2em}

\subcaptionbox{Selected steps in computation of the logical form of the object relative clause. We omit tense information for space.\label{fig:objrelclause}}[\textwidth]{
\begin{tabular}{>{\centering\let\newline\\\arraybackslash\hspace{0pt}}p{6cm}>{\centering\let\newline\\\arraybackslash\hspace{0pt}}p{4cm}>{\centering\let\newline\\\arraybackslash\hspace{0pt}}p{5.5cm}}
\tll \obj{a mouse chased}\trr & \tll\obj{cat}\trr & \tll \obj{cat a mouse chased}\trr \newline = \tll acl\trr \(\big(\)\tll\obj{cat}\trr\(\big)\big(\)\tll \obj{a mouse chased}\trr\(\big)\) \\
\((e_{oj}(st))\) & \((et)\) & \((et)\)\\
\(\lambda x. \lambda e.
\begin{array}{|c|}\hline
y\\\hline
\textup{chase}(e)\\
\textup{Theme}(e,x)\\
\textup{Agent}(e,y)\\
\textup{mouse}(y)
\\\hline\end{array}
\) & \(
\lambda x.
\begin{array}{|c|}\hline
\\\hline
\textup{cat}(x)
\\\hline\end{array}
\) & \(
\lambda x.
\begin{array}{|c|}\hline
y\ e\\\hline
\textup{cat}(x)\\
\textup{chase}(e)\\
\textup{Theme}(e,x)\\
\textup{Agent}(e,y)\\
\textup{mouse}(y)
\\\hline\end{array}
\)
\end{tabular}
}

\caption{Examples of computation of logical forms for both subject and object relative clauses.}
\label{fig:relativeclause}
\end{figure*}

\begin{figure*}
\centering

\subcaptionbox{The Universal Dependencies syntax of a sentence with quantifier scope ambiguity.
\label{fig:scopeambigsyntax}}[\textwidth]{
\begin{dependency}
\begin{deptext}
Every \& cat \& chased \& a \& mouse \\
DET \& NOUN \& VERB \& DET \& NOUN\\
\end{deptext}
\depedge{2}{1}{det}
\depedge{5}{4}{det}
\depedge{3}{2}{nsubj}
\depedge{3}{5}{obj}
\end{dependency}}

\vspace{2em}

\subcaptionbox{Semantic types and denotations of \obj{every cat}, \obj{a mouse}, and \obj{chased}.\label{fig:everycat}}[\textwidth]{
\begin{tabular}{rll}
\tll \obj{every cat}\trr:& \(((et)t)\) & \(\lambda F.[ \neg [\exists x.[ \textup{cat}(x) \wedge \neg [F(x)]]]]\)\\
\tll \obj{a mouse}\trr:  & \(((et)t)\) & \(\lambda F. \exists y.[ \textup{mouse}(y) \wedge F(y)] \)\\
\tll \obj{chased}\trr:  & \((e_{oj}(e_{sj}(st)))\) & \(\lambda y. \lambda x. \lambda e.[\textup{chase}(e) \wedge \textup{Theme}(e,y) \wedge \textup{Agent}(e,x)]\)
\end{tabular}
}

\vspace{2em}

\subcaptionbox{Semantic types and denotations of ``nsubj'' and ``obj'' used to compute surface quantifier scope of the sentence in Figure \ref{fig:scopeambigsyntax}.}[\textwidth]{
\begin{tabular}{rll}
\tll nsubj\trr: & \(((e_{sj}(st))\ (((et)t)\ (st)))\) & \(\lambda F. \lambda G. \lambda e. [G(\lambda x. [F(x,e)])]\)\\
\tll obj\trr: & \(((e_{oj}(e_{sj}(st)))\ (((et)t)\ (e_{sj}(st))))\) & \(\lambda F. \lambda G. \lambda x. \lambda e. [G(\lambda y. [F(y,x,e)])]\)
\end{tabular}
}

\vspace{2em}

\subcaptionbox{Selected steps in computation of the logical form of  the sentence in Figure \ref{fig:scopeambigsyntax} with surface quantifier scope.}[\textwidth]{
\small \begin{tabular}{>{\centering\let\newline\\\arraybackslash\hspace{0pt}}p{6cm}>{\centering\let\newline\\\arraybackslash\hspace{0pt}}p{10cm}}
\tll\obj{chased a mouse}\trr\newline = 
\(\llbracket\textup{obj}\rrbracket(\llbracket\textup{\obj{chased}}\rrbracket)(\llbracket\textup{\obj{a mouse}}\rrbracket)\) & \tll\obj{every cat chased a mouse}\trr\newline = \(
\llbracket\textup{nsubj}\rrbracket\Big(\llbracket\textup{obj}\rrbracket(\llbracket\textup{\obj{chased}}\rrbracket)(\llbracket\textup{\obj{a mouse}}\rrbracket)\Big)(\llbracket\textup{\obj{every cat}}\rrbracket)\)\\
\((e_{sj}(st))\) & \((st)\) \\
\(\lambda x. \lambda e.
\begin{array}{|c|}\hline
y\\\hline
\textup{chase}(e)\\
\textup{Theme}(e,y)\\
\textup{Agent}(e,x)\\
\textup{mouse}(y)
\\\hline\end{array}\) &
\(\lambda e.
\begin{array}{|c|}\hline
\\\hline
\\[-0.5em]\neg\ \begin{array}{|c|}\hline
x\\\hline
\textup{cat}(x)\\
\neg\ \begin{array}{|c|}\hline
y\\\hline
\textup{chase}(e)\\
\textup{Theme}(e,y)\\
\textup{Agent}(e,x)\\
\textup{mouse}(y)
\\\hline\end{array}
\\\hline\end{array}
\\\hline\end{array}\)
\end{tabular}}

\vspace{2em}

\subcaptionbox{Semantic types and denotations of ``nsubj'' and ``obj'' used to compute inverse quantifier scope of the sentence in Figure \ref{fig:scopeambigsyntax}.}[\textwidth]{
\begin{tabular}{rll}
\tll nsubj\trr: & \(((e_{oj}(e_{sj}(st)))\ (((et)t)\ (e_{oj}(st))))\) & \(\lambda F. \lambda G. \lambda y. \lambda e. [G(\lambda x. [F(y,x,e)])]\) \\
 \tll obj\trr: & \(((e_{oj}(st))\ (((et)t)\ (st)))\) & \(\lambda F. \lambda G. \lambda e. [G(\lambda y. [F(y,e)])]\)
\end{tabular}
}

\vspace{2em}

\subcaptionbox{Selected steps in computation of the logical form of the sentence in Figure \ref{fig:scopeambigsyntax} with inverse quantifier scope.}[\textwidth]{
\small \begin{tabular}{>{\centering\let\newline\\\arraybackslash\hspace{0pt}}p{6cm}>{\centering\let\newline\\\arraybackslash\hspace{0pt}}p{10cm}}
\tll\obj{every cat chased}\trr\newline = 
\(\llbracket\textup{nsubj}\rrbracket(\llbracket\textup{\obj{chased}}\rrbracket)(\llbracket\textup{\obj{every cat}}\rrbracket)\) & \tll\obj{every cat chased a mouse}\trr\newline = \(\llbracket\textup{obj}\rrbracket\Big(\llbracket\textup{nsubj}\rrbracket(\llbracket\textup{\obj{chased}}\rrbracket)(\llbracket\textup{\obj{every cat}}\rrbracket)\Big)(\llbracket\textup{\obj{a mouse}}\rrbracket)\)\\
\((e_{oj}(st))\) & \((st)\) \\
\(\lambda y. \lambda e.
\begin{array}{|c|}\hline
\\\hline
\\[-0.5em]\neg\ \begin{array}{|c|}\hline
x\\\hline
\textup{cat}(x)\\
\neg\ \begin{array}{|c|}\hline
\\\hline
\textup{chase}(e)\\
\textup{Theme}(e,y)\\
\textup{Agent}(e,x)
\\\hline\end{array}
\\\hline\end{array}
\\\hline\end{array}\) &
\(\lambda e. \begin{array}{|c|}\hline
y\\\hline
\textup{mouse}(y)\\
\neg\ \begin{array}{|c|}\hline
x\\\hline
\textup{cat}(x)\\
\neg\ \begin{array}{|c|}\hline
\\\hline
\textup{chase}(e)\\
\textup{Theme}(e,y)\\
\textup{Agent}(e,x)
\\\hline\end{array}
\\\hline\end{array}
\\\hline\end{array}\)
\end{tabular}}

\caption{Computations of logical form for both possible quantifier scopes in the sentence \obj{Every cat chased a mouse.} We omit tense information for space.}
\label{fig:scopeambiguity}
\end{figure*}

\section{Theoretical Choices in Constructing Lexicon}
\label{app:theorysection}

In this section, we outline reasoning from formal linguistics which motivates choices we have made in crafting our lexicon.

\subsection{Nominal Domain}
\label{app:nounmeanings}

We take nouns on their own to denote one-place predicates of an entity, that is, to have type \((et)\). This makes it simple for adjectives to modify nouns: The denotation of the adjective relation ``amod'' maps two predicates to their logical conjunction.

We take all phrases headed by nouns to have type \(((et)t)\) and to denote generalized quantifiers over predicates of one entity, as proposed by \citet{barwiseandcooper}. For example, the meaning of \emph{every cat} (in Figure \ref{fig:everycat}) takes a one-place predicate as input, and outputs a DRS asserting that this predicate holds of every cat. The meaning of \emph{a mouse} (in the same figure) takes a predicate as input, and outputs a DRS asserting that there exists a mouse that this predicate holds of. The meaning of \emph{the big red dog} takes a predicate as input, and outputs a DRS presupposing the existence of a big red dog, and asserting that this predicate holds of that dog (the output of which process is part of Figure \ref{fig:bigreddogmeaning}). As shown by these examples, and as observed by Barwise and Cooper, the uniform semantic type  \(((et)t)\) allows us compute meanings for noun phrases with a variety of determiners. Moreover, having a uniform semantic type for noun phrases simplifies the denotations of relations which take noun phrases as dependents, such as \tll nsubj\trr. Finally, as we compute in Appendix \ref{app:workedexamples}, treating noun phrase meanings as generalized quantifiers allows us to derive every possible scope relationship in a sentence with multiple quantifiers. Thus, \(((et)t)\) is a desirable semantic type for a noun phrase.

This means that no noun can be a complete noun phrase on its own, as nouns have type \((et)\) while noun phrases have type \(((et)t)\). When a noun appears on its own, we therefore depart from the Universal Dependencies standard and assume it has a silent determiner. If it is a proper noun, it has a silent definite determiner (synonymous with \obj{the} in English), and if it is a common noun, it has a silent indefinite determiner (synonymous with \obj{a} in English).

There are several reasons for this approach. First, proper nouns have a definite description meaning. They introduce a presupposition that a referent with the appropriate name exists. Moreover, some proper nouns, such as \obj{The Netherlands}, transparently have a definite determiner as part of their structure, and others, like \obj{Belgium}, do not, and this distinction does not affect their semantic definiteness. This is even more apparent cross-linguistically. In some languages, such as Modern Greek, it is common to use an overt definite determiner when using a person's name \cite{papaloizos}, while in other languages it is not, but this too makes no difference to the semantic definiteness of the proper noun. So, proper nouns must contain the meaning of a definite determiner.

Moreover, we should separate the part of a proper noun which introduces the referent's name from the part which supplies the definiteness. This is for two reasons. First, a proper noun may be modified by an adjective, in which case the meaning of the adjective is included in the presupposition introduced by the proper noun. For example, the sentence \obj{Young Peter came to visit today} presupposes that a person exists who is named Peter and who is young. We therefore wish the adjective to combine with the proper noun before the definite determiner does. Secondly, when a proper noun has an overt determiner, the definiteness in its meaning can vanish. For example, the sentence \obj{Every Peter in my class has a pet poodle} does not presuppose the existence of anyone named Peter. Consequently, it is beneficial to assume that the definiteness in a proper noun lacking an overt determiner really comes from a silent definite determiner, separate from the noun itself.

The case of common nouns is similar. A common noun in a particular context can have the same meaning across languages, even if one language uses no determiner and another uses an indefinite determiner. For example, the sentences \emph{John became a lawyer} in English and \emph{John est devenu avocat} in French have the same meaning, but the English sentence includes an indefinite determiner while the French one does not. Just as in the proper noun case, only the presence of an overt determiner can change the noun's meaning to anything other than an indefinite description.

For these reasons, despite the Universal Dependencies guidelines against null elements, we posit the existence of silent determiners on any common or proper noun that lacks an overt determiner.

\subsection{Verbal Domain}
\label{app:verbmeanings}

We take the denotation of a verb to be a relation between at least one entity, an event representing the verb's action, and potentially a non-entity argument which may be an entire clause or an event (for certain types of embedding verbs).

We also assume that the lexical entry for a verb includes its semantic valency, as well as the semantic argument role filled by each of its arguments. For example, a transitive verb's denotation is a function with two arguments of type \(e\) and one argument of type \(s\), returning a DRS stating that the entity arguments are the participants in the action denoted by the event argument. We make the additional assumption that verbs' denotations specify the syntactic position of each entity argument they take, by means of our syntactically-labelled subtypes of \(e\). For example, \obj{sleep} has type \((e_{sj}(st))\), taking one entity subject, and \obj{say} has type \((t(e_{sj}(st)))\), taking one clausal argument and one entity subject.

The denotations of verbs are therefore quite simple in structure. They can be so simple because the denotations of relations do the work of saturating their arguments. So, the inputs to the subject relation \(\llbracket \textup{nsubj} \rrbracket\) may be a verb denotation of type \((e_{sj}(st))\) and a noun phrase denotation of type \(((et)t)\), in which case the function \(\llbracket \textup{nsubj} \rrbracket\) outputs a phrase of type \((st)\). The ``root'' relation denotes a one-place function of type \(((st)\ t)\) and has the effect of existentially closing the open event argument of a verb.

By tagging verb arguments with their syntactic relationship to the verb, we allow for flexibility in the binarization order of phrases headed by verbs, without sacrificing precision as to which argument is which. This flexibility has two benefits. First, when combined with the denotations of noun phrases as generalized quantifiers (described in Section \ref{app:nounmeanings}), this allows for the computation of both surface and inverse scope readings in transitive sentences with quantified subject and object. Figure \ref{fig:scopeambiguity} in Appendix \ref{app:workedexamples} shows an example.

Second, in a situation with one verb argument moved or missing, such as a relative clause or participle, the denotations of embedding relations like ``acl'' (adjectival clause, used for relative clauses) or ``xcomp'' (embedded clause missing an argument) can reliably saturate the correct argument of the embedded verb. An example is shown in Figure \ref{fig:relativeclause} in Appendix \ref{app:workedexamples}. In Figure \ref{fig:subjrelclause} we correctly interpret the cat as the Agent of the chasing event, whereas in Figure \ref{fig:objrelclause} we correctly interpret the cat as the Theme of the chasing event.

\section{Comparison with CCG}
\label{ccgappendix}

As a joint syntax and semantics, UD with UD-TC invites comparison with Combinatory Categorial Grammar \cite[CCG;][]{ccg}, which is widely used for computational applications. Both express word meanings as semantically-typed functions. One difference is that in CCG, word meanings are combined by different \emph{combinators}, including functional application, but also including function composition and type raising. These are vital to CCG meaning computations in many contexts, such as relative clauses. In UD-TC, meanings are always combined by functional application, and due to the expressivity of relation meanings, we still capture a variety of phenomena. Another difference is that CCG seeks not only to compute meanings, but also to account for the grammaticality or ungrammaticality of certain word orders. By contrast, UD-TC ignores word order, which lets it generalize more easily to languages with different word orders.

\end{document}